\documentclass[journal, twoside]{IEEEtran}

\usepackage[pdftex]{graphicx}
\ifCLASSOPTIONcompsoc
  \usepackage[caption=false,font=normalsize,labelfont=sf,textfont=sf]{subfig}
\else
  \usepackage[caption=false,font=footnotesize]{subfig}
\fi
\usepackage{comment}
\usepackage{mathtools}
\usepackage{amsfonts}
\usepackage{bm}
\usepackage{siunitx}
\usepackage{textcomp}
\usepackage{hyperref}

\usepackage{tikz}
\usetikzlibrary{calc}
\usetikzlibrary{shapes.geometric, arrows, fit}
\usepackage{varwidth}
\usetikzlibrary{positioning}
\usetikzlibrary{backgrounds}
\usetikzlibrary{decorations.pathreplacing}

\usepackage{pgfplots,filecontents}

\DeclareMathOperator{\vect}{vec}

\DeclareRobustCommand\changed{\textcolor{black}}

\newcommand{\R}{\mathbb{R}}

\newcommand{\D}{\mathcal{D}}

\definecolor{navy}{RGB}{0,0,137}
\definecolor{tealDeer}{RGB}{148,232,180}
\definecolor{dodgerBlue}{RGB}{18,161,255}
\definecolor{citrine}{RGB}{230, 194, 8}
\definecolor{violet}{RGB}{112,5,164}
\definecolor{navyPurple}{RGB}{172,86,253}
\definecolor{heliotrope}{RGB}{236,93,253}
\definecolor{pink}{RGB}{250,38,160}

\ifCLASSOPTIONpeerreview
  \usepackage{setspace}
\fi

\begin{document}

\author{Italos~Estilon~de~Souza,~and~Alexandre~Xavier~Falc\~{a}o}

%!TEX root = arxiv.tex

\title{Learning CNN filters from user-drawn image markers for coconut-tree image classification}

% make the title area
\maketitle

\begin{abstract}
Identifying species of trees in aerial images is essential for land-use classification, plantation monitoring, and impact assessment of natural disasters. The manual identification of trees in aerial images is tedious, costly, and error-prone, so automatic classification methods are necessary. Convolutional Neural Network (CNN) models have well succeeded in image classification applications from different domains. However, CNN models usually require intensive manual annotation to create large training sets. One may conceptually divide a CNN into convolutional layers for feature extraction and fully connected layers for feature space reduction and classification. We present a method that needs a minimal set of user-selected images to train the CNN's feature extractor, reducing the number of required images to train the fully connected layers. The method learns the filters of each convolutional layer from user-drawn markers in image regions that discriminate classes, allowing better user control and understanding of the training process. It does not rely on optimization based on backpropagation, and we demonstrate its advantages on the binary classification of coconut-tree aerial images against one of the most popular CNN models.
\end{abstract}


\section{Introduction}

\IEEEPARstart{\changed{D}}{\changed{eep}} learning has proven to be applicable to different tasks, from image classification to data synthesis \cite{goodfellow2016deep}. In remote sensing, the applications may involve segmentation of terrain images \cite{kemker2018algorithms, kampffmeyer2016semantic}, building identification \cite{xu2018building, lu2018detecting, liu2018multilevel}, and deforestation monitoring \cite{bragilevsky2017deep}, for instance. In this work, we are interested in identifying species of trees from aerial images. The topic is important for land-use classification, plantation monitoring, and damage assessment of natural disasters. As the plantations can span vast areas, the manual identification of each tree is costly, tedious, and error-prone, and so automatic classification methods are necessary. 

Classification of tree species in aerial images has been actively investigated~\cite{fassnacht2016review}. In \cite{puttemans2018comparing} and~\cite{aparna2018cnn}, the authors present automatic solutions to detect coconut trees based on convolutional neural network (CNN) models. Despite these recent advances, CNN models usually require considerable human effort in image annotation to create large training sets. Vargas-Muñoz et al.~\cite{8899005} propose an active learning approach to mitigate the problem. The method explores data projection techniques to allow simultaneous annotation of multiple regions as having or not coconut trees. It then uses a CNN to identify the candidate regions with coconut trees. We adopt another alternative -- the design of simplified CNN models from small training sets with user interaction, being the user knowledgeable about CNN modeling and the application domain.

The design of CNN models without a human as part of the training loop leaves several questions unanswered: (1) How to find a useful and simplified model for a given classification problem? (2) How to train that model from a minimum number of annotated images?  (3) Can the user explain the decisions of the model? (4) Can the model improve from label corrections? The first question requires human knowledge about CNN modeling and the problem of interest. The second one requires to reduce the human effort to train a CNN model. The third issue is related to human understanding. It may explore visual analytics to explain the model's decisions and to guide the user in the design of the model~\cite{RauberInfoVis2018,GarciaIJCNN2019,SpinnerTVCG2020}. The fourth question is also essential during training, and it is related to user control over the process. They all lead to the importance of involving human experts during the machine learning process. \changed{Explainable artificial intelligence (XAI) has appeared to address (3)~\cite{rauber2016visualizing,arrietaInfoFusion2020}. However, the importance of human participation during the training process has called little attention~\cite{AngelovNN2020}. Exceptions rely on visual analytics~\cite{RauberInfoVis2018,GarciaIJCNN2019,SpinnerTVCG2020}, and we present here a first work in which the user has direct influence over the quality of the filters.}

First, we conceptually divide a CNN into convolutional layers for feature extraction and fully connected layers for feature space reduction and classification. Each convolutional layer contains a filter bank, an activation function, and alternative operations (e.g., pooling, batch normalization).  As the number of convolutional layers increases (deeper is the model), higher is the number of annotated images required to train the model by backpropagation. 

To reduce the need for large annotated training sets by exploiting the user knowledge, we present a method, called \changed{\emph{Feature Learning from Image Markers} (FLIM)}, that needs a minimal set of images to learn the filters of each convolutional layer. The user selects the number of convolutional layers, their filter sizes, \changed{a few images from a training set}, and draws markers in image regions that best discriminate the classes. The user takes a couple of minutes to select images and draw markers in those images. \changed{By clustering patches extracted from marker pixels at the input of each convolutional layer, FLIM finds filters that enhance those regions at the convolutional layers' output. The remaining training images can then be used to train the fully connected layers or another classifier.} The CNN architecture could also be optimized \changed{with the support of} visual analytics~\cite{SpinnerTVCG2020}, but we have not explored this option yet. \changed{In interactive semantic segmentation, user-selected clicks can affect a part of the network to improve segmentation, but the filters are not designed from those markers~\cite{SofiiukCVPR2020}.}

We demonstrate the advantages of this interactive technique on binary classification of coconut-tree images against  VGG-16~\cite{simonyan2014very}. First, the user can better understand and control the training process by observing the effectiveness of marker selection.  Second, very few images per class (e.g., less than five) seem to be enough. Third, by eliminating the optimization of the convolutional layers by backpropagation, FLIM can reduce the number of annotated images to train the fully connected layers. Fourth, FLIM is application-independent, and so it might be useful for other image classification problems.

\section{FLIM: Feature Learning from Image Markers}
\label{sec:method}
  
\changed{A CNN model should enhance image regions that best discriminate classes through a sequence of convolutional layers. This section shows how identifying such regions by a simple user action can make that sequence an effective feature extractor. The method is named \emph{Feature Learning from Image Markers} (FLIM). In the feature space that results from convolutional layers,} it is common to assume that the classes are piecewise linearly separable and apply fully connected layers -- Multi-Layer Perceptron (MLP) -- for feature space reduction and classification. The training of a CNN usually relies on weight optimization using backpropagation, which requires a higher number of annotated images as deep as the model is. In FLIM, we eliminate the need for backpropagation to train convolutional layers by finding a set ${\cal F}$ of filters that can \changed{enhance} discriminant regions from all classes.

The user selects a few \changed{images from a training set} to compose a very small dataset $\D$ and draws labeled markers in image regions that best discriminate the classes. The convolution between an image $I$ and a filter $F\in {\cal F}$ \changed{ enhances} regions whose local patterns are detected by $F$. We wish then to estimate the weights of $F$ such that those local patterns are characteristic of one given class. For sake of clarity, one must interpret the convolution operation at a pixel $p$ of $I$ as the inner product $\vect(P_I(p)) \cdot \vect(F)$, where $P_I(p)$ is a $k\times k$ patch with $m$ \changed{bands} around $p$, $F$ is a filter with shape $k\times k \times m$, and $\vect$ is the vectorization operation. A filter $F$ can discriminate a class $i \in \{1,2,\ldots,c\}$ among $c$ classes when it generates positive results for local patterns $\vect(P_I(p))$ of class $i$ and negative results for patterns from other classes. That is, $\vect(F)$ is the normal vector of a hyperplane in $\R^{k\times k\times m}$, which detects the patterns of class $i$ at its positive side. We wish to estimate ${\cal F}$ such that its filters will detect patterns from all classes in different positions $p$. 

\begin{figure}[!ht]
  \centering
  \subfloat[\label{fig:ex-groups}]
  {\includegraphics[width=.35\linewidth]{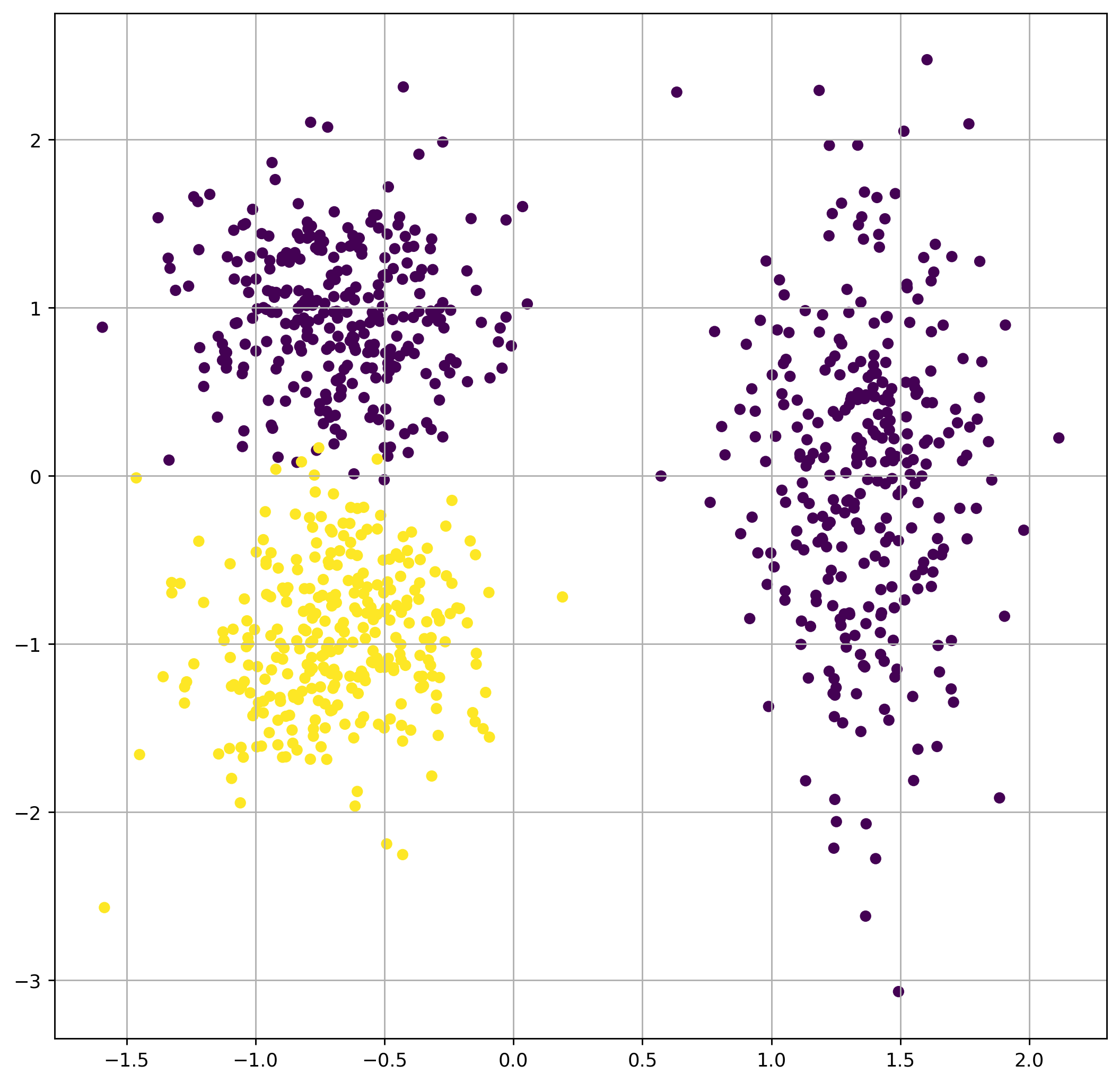}}
  ~
  \subfloat[\label{fig:ex-groups-after}]{\includegraphics[width=.45\linewidth]{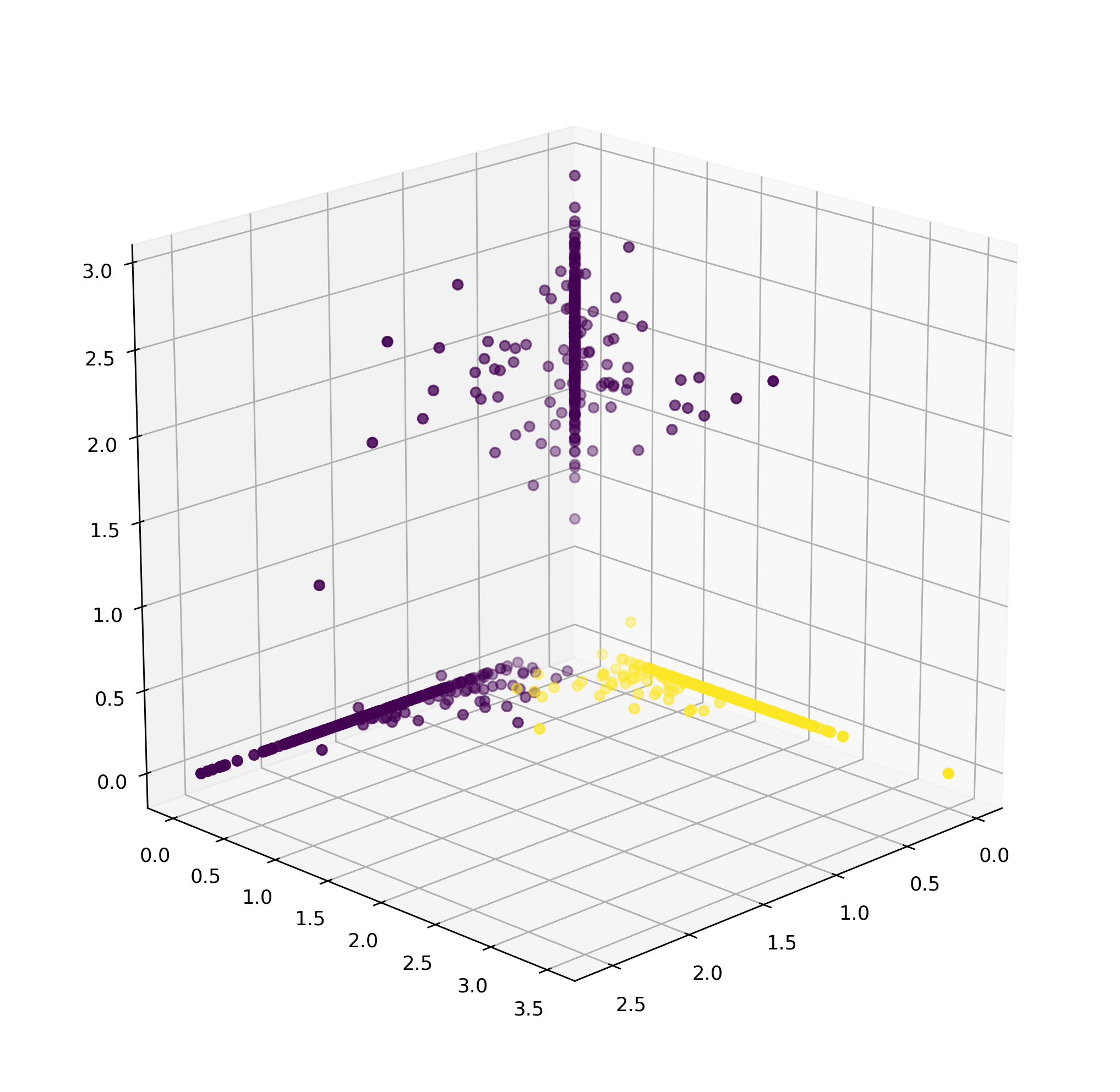}
  }
  \caption{(a) A 2D feature space with color-coded samples from two classes distributed in three groups. (b) After marker-based batch normalization and convolution with filters obtained from each group, the classes can be more easily detected by two filters in a new 3D feature space.}
  \label{fig:filter}
\end{figure}

For a problem with $c$ classes where $\lambda(p)=i\in \{1,2,\ldots,c\}$ is the label of a marker pixel $p$ from class $i$, let ${\cal M}_I$ be a set of marker pixels drawn in image $I\in {\cal D}$ and $P_I(p)$ be the respective patch around a pixel $p\in {\cal M}_I$. Let ${\cal P}_i$ be the set of all patches around marker pixels from all images $I\in {\cal D}$, with representative examples  $\vect(P_I(p))$ from class $i\in \{1,2,\ldots,c\}$. 
\begin{eqnarray}
{\cal P}_i & = & \bigcup_{I\in {\cal D},p\in {\cal M}_I, \lambda(p) = i}{P_I(p)}.
\label{eq:pset}
\end{eqnarray}
A clustering operation on each set ${\cal P}_i$, $i\in [1,c]$, guarantees groups with similar local patterns for each class. The groups must be shifted to the origin of $\R^{k\times k\times m}$ by subtracting the mean value of all patches, set $\mathcal{P} = \bigcup_{i \in \{1, 2, \ldots, c\}}{\mathcal{P}_i}$. Additionally, the standard deviation of all patches is computed and used for standardization. This operation allows batch normalization of image sets using the mean and standard deviation of the patches from the image markers -- i.e.,  \emph{a marker-based batch normalization}. The centroid of each cluster defines the weights of each filter $F\in {\cal F}$, and we force norm $\|\vect(F)\|=1$ to avoid preferences among them. The centralization for filter definition is paramount to eliminate activations in regions from other classes, whose patterns fall in the negative side of the hyperplane with normal $\vect(F)$. Figure~\ref{fig:filter} illustrates three groups with color-coded samples from two classes in a hypothetical 2D feature space. The marker-based batch normalization and convolution with filters obtained as centroids of the three groups in Figure~\ref{fig:filter}a creates a new 3D feature space (Figure~\ref{fig:filter}b), in which the classes can be more easily detected by two filters in distinct activation \changed{bands}.

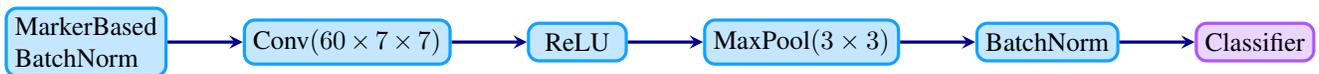
\begin{figure*}
  \begin{center}
  
    \tikzstyle{steps} = [rectangle, rounded corners, minimum width=1.3cm, minimum height=0.5cm,text centered, draw=dodgerBlue, very thick, fill=dodgerBlue!25, execute at begin node={\begin{varwidth}{2.5cm}},
      execute at end node={\end{varwidth}}]
  
      \tikzstyle{myarrow} = [navy, very thick]
  
      \tikzstyle{arrow} = [thick,->,>=stealth]

  \begin{tikzpicture}
    
    \node (mbn) [steps] {MarkerBased \\ BatchNorm};

    \node (conv) [steps, right= of mbn] {Conv$(60 \times 7 \times 7)$};
       
    \node (relu) [steps, right= of conv] {ReLU}; 

    \node (pool) [steps, right=of relu] {MaxPool$(3\times3)$};

    \node (bn) [steps, right=of pool] {BatchNorm};

    \node (clf) [steps, fill=navyPurple!25, draw=navyPurple, right= of bn] {Classifier};

    %-----------------------arrows---------------

    \draw [arrow, myarrow] (mbn) -- (conv);
    \draw [arrow, myarrow] (conv) -- (relu);
    \draw [arrow, myarrow] (relu) -- (pool);
    \draw [arrow, myarrow] (pool) -- (bn);
    \draw [arrow, myarrow] (bn) -- (clf);

  \end{tikzpicture}
  \end{center}
  \caption{Example of a network with a single convolutional layer followed by a classifier, as projected by FLIM and\changed{,} used for the experiments in this work.}
  \label{fig:arch}
\end{figure*}

Each convolutional layer is trained individually, one layer at a time, to find its filter set ${\cal F}$. The number of filters per layer depends on the clustering technique. We use $K$-means, and then the user must specify the number $K=|{\cal F}|$ of filters. After the convolution operation, we apply the ReLU function to eliminate negative activations and the max-pooling operation to aggregate local information. Note that, apart from the initial marker selection, the training process is automatic. We have preserved the convolutional layers' output dimensions and the marker pixels so we can find the filters of layer $L+1$ from the output of layer $L$. Figure~\ref{fig:arch} shows one example of a simple network projected by FLIM for the experiments of the next section. 

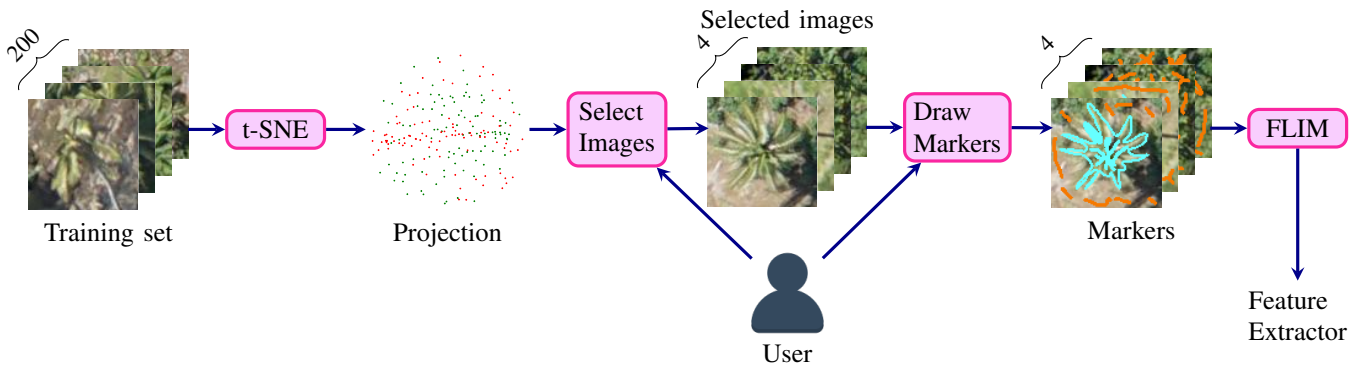
\begin{figure*}
  \begin{center}
    
    \tikzstyle{steps} = [rectangle, rounded corners, minimum width=1.3cm, minimum height=0.5cm,text centered, draw=pink, very thick, fill=heliotrope!30, execute at begin node={\begin{varwidth}{2.5cm}},
    execute at end node={\end{varwidth}}]
    \tikzstyle{image} = [minimum width=1.3cm, minimum height=0.7cm,text centered, draw=white,fill=white!30, inner sep=0pt]

    \tikzstyle{myarrow} = [navy, very thick]

    \tikzstyle{arrow} = [thick,->,>=stealth]

    \begin{tikzpicture}
    %images----------------------------
     \node (img1) [image] {\includegraphics[width=.08\textwidth]{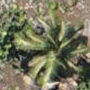}};
     \node (img2) [image, below left=-5em of img1] {\includegraphics[width=.08\textwidth]{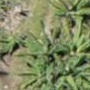}};
     \node (img3) [image, below left=-5em of img2] {\includegraphics[width=.08\textwidth]{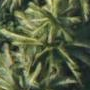}};
     \node (img4) [image, below left=-5em of img3] {\includegraphics[width=.08\textwidth]{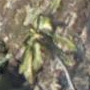}};

     \node (imgs) [fit={($(img1.north west)+(-0.64,0.0)$) ($(img4.south east)+(0.64,-0.0)$)}, inner sep=0pt, draw=none, rounded corners, very thick, label = {[label distance = 0.1em]below:{Training set}}] {};
     %----------------------------------

     \node (tsne) [steps, right=1.4em of imgs] {t-SNE};

     \node (vis1) [image, right=1.5em of tsne, outer sep=0pt, label = {[label distance = 0.1em]below:{Projection}}] {\includegraphics[width=.12\textwidth]{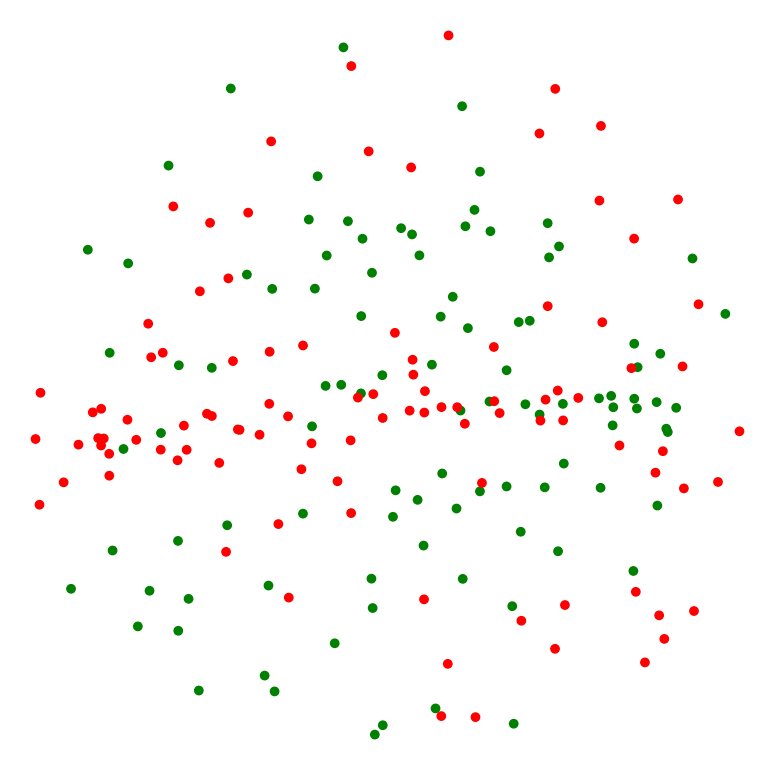}};

     \node (select) [steps, right=1.4em of vis1] {Select\\ Images};
    
    %images----------------------------
    \node (simg1) [image, above right=-2.5em and 3.4em of select] {\includegraphics[width=.08\textwidth]{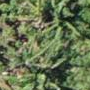}};
    \node (simg2) [image, below left=-5em of simg1] {\includegraphics[width=.08\textwidth]{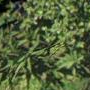}};
    \node (simg3) [image, below left=-5em of simg2] {\includegraphics[width=.08\textwidth]{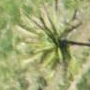}};
    \node (simg4) [image, below left=-5em of simg3] {\includegraphics[width=.08\textwidth]{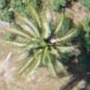}};

    \node (simgs) [fit={($(simg1.north west)+(-0.64,0.0)$) ($(simg4.south east)+(0.64,-0.0)$)}, inner sep=0pt, draw=none, label = {[label distance = 0.15em]above:{Selected images}}] {};
    %----------------------------------

     \node (put) [steps, right=1.4em of simgs] {Draw\\Markers};

     %images----------------------------
    \node (mimg1) [image, above right=-2.5em and 3.4em of put] {\includegraphics[width=.08\textwidth]{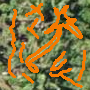}};
    \node (mimg2) [image, below left=-5em of mimg1] {\includegraphics[width=.08\textwidth]{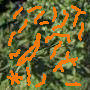}};
    \node (mimg3) [image, below left=-5em of mimg2] {\includegraphics[width=.08\textwidth]{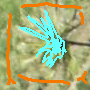}};
    \node (mimg4) [image, below left=-5em of mimg3] {\includegraphics[width=.08\textwidth]{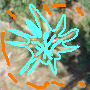}};

    \node (mimgs) [fit={($(mimg1.north west)+(-0.64,0.0)$) ($(mimg4.south east)+(0.64,-0.0)$)}, inner sep=0pt, draw=none, label = {[label distance = 0.1em]below:{Markers}}] {};

    \node (flim) [steps, right=1.4em of mimgs] {FLIM};

    \node (ft) [steps, below=5em of flim, fill=white, draw=none] {Feature\\Extractor};

    \node (user) [image, below=1.9em of simgs, outer sep = -.5em, label = {[label distance = 0.3em]below:{User}}] {\includegraphics[width=.07\textwidth]{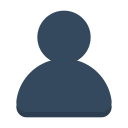}};

     %--------------------------------------------------------------arrows
     \draw [arrow, myarrow] (imgs) -- (tsne);
     \draw [arrow, myarrow] (tsne) -- (vis1);
     \draw [arrow, myarrow] (vis1) -- (select);
     \draw [arrow, myarrow] (select) -- (simgs);
     \draw [arrow, myarrow] (simgs) -- (put);
     \draw [arrow, myarrow] (put) -- (mimgs);
     \draw [arrow, myarrow] (mimgs) -- (flim);
     \draw [arrow, myarrow] (flim) -- (ft);
     \draw [arrow, myarrow] (user) -- (select);
     \draw [arrow, myarrow] (user) -- (put);

     \draw [decorate,decoration={brace,amplitude=4pt,raise=5pt}]
(img4.north west) -- (img1.north west) node [black,midway,xshift=-0.4cm, yshift=0.4cm, rotate=45] {\small 200};

\draw [decorate,decoration={brace,amplitude=4pt,raise=5pt}]
(simg4.north west) -- (simg1.north west) node [black,midway,xshift=-0.4cm, yshift=0.4cm, rotate=45] {\small 4};

\draw [decorate,decoration={brace,amplitude=4pt,raise=5pt}]
(mimg4.north west) -- (mimg1.north west) node [black,midway,xshift=-0.4cm, yshift=0.4cm, rotate=45] {\small 4};

    \end{tikzpicture}
  \end{center}
  \caption{\changed{Pipeline for training a feature extractor using FLIM. In the projection, green points are images with coconut trees, and red points are images with no coconut trees. The user selects a few images (e.g., 4) from the projection. The user draws orange markers on background pixels and cyan markers on coconut-tree pixels for image feature learning by FLIM.}}
  \label{fig:pipeline-xtractor}
\end{figure*}

\section{Experiments}

We use a dataset from~\cite{8899005}, which contains regions of aerial images with and without coconut trees from the Kingdom of Tonga, as acquired in October 2017~\footnote{The images are available in \url{blog.werobotics.org/2018/01/11/open-ai-challenge-2}.}. Each region is classified as containing one or none coconut tree, but part of coconut trees from adjacent regions might appear near the region's border. The Humanitarian OpenStreetMap community annotated the regions. The dataset consists of 13587 regions with $90 \times 90$ pixels \changed{and spatial resolution of $8$\si{\centi\metre}}, being 10268 and 3319 regions annotated as containing and not containing coconut trees, respectively. \changed{The region images contain three bands: red, green, and blue. We convert them to the CIE L*a*b* color space and normalize each band within $[0,1]$}.

\changed{The experiments used an Intel(R) Xeon(R) CPU E5-2620 v4 @ 2.10GHz CPU and a Titan X with 12 GB of memory. They have been designed to evaluate the following hypotheses about FLIM.
\begin{enumerate}
  \item[H1]  It can produce competitive feature extractors from user-drawn markers in a few images selected per class from a training set. 
  \item[H2]  For a given network architecture, FLIM can provide better classification results than those obtained by training the same architecture with backpropagation.
  \item[H3] Its feature extractor remains competitive independently of the training set size. 
  \item[H4] User-drawn markers on a higher number of selected images can improve feature extraction.
\end{enumerate}
}

\changed{For these experiments, we have selected the following models that combine feature extraction and classification methods.
 \begin{itemize}
\item FLIM+SVM: Image classification is performed by support vector machines (SVM) from FLIM-based features.  
\item FLIM+MLP: Image classification is performed by \changed{Multi-Layer Perceptron} (MLP) from FLIM-based features.  
\item FLIM-FT+MLP: The architecture in FLIM+MLP is fine-tuned by backpropagation. 
\item CL+MLP: The architecture in FLIM+MLP is trained from scratch by backpropagation.
\item VGG: VGG-16~\cite{simonyan2014very} is trained from scratch.
\item VGG-FT: VGG-16 is pre-trained on ImageNet and fine-tuned on the coconut-tree training set.
\item VGG-FE+SVM: Image features are obtained from the last convolutional layer of VGG-FT.
\end{itemize}
}

\begin{figure}[!t]
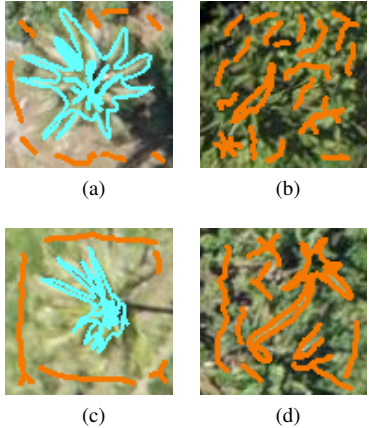

    \centering
    \subfloat[\label{fig:markers1}]{\includegraphics[width=.25\linewidth]{figs/markers1}
    }
    ~
    \subfloat[\label{fig:markers2}]{\includegraphics[width=.25\linewidth]{figs/markers2}
    }
    \\
    \subfloat[\label{fig:markers3}]{\includegraphics[width=.25\linewidth]{figs/markers3}
    }
    ~
    \subfloat[\label{fig:markers4}]{\includegraphics[width=.25\linewidth]{figs/markers4}
    }
    \caption{\changed{Markers used for training, cyan markers on coconut-tree pixels and orange markers on background pixels.}}
    \label{fig:markers}
\end{figure}

\changed{First, we evaluated H1 and H2 by randomly selecting 200 images for the training set, 2000 images for a validation set, and the remaining 11387 images were left in the testing set. The validation set was used only once to choose empirically the architecture of the network shown in Figure~\ref{fig:arch}, and to choose the training hyperparameters. The convolutional layer has $60$ filters with dimension $7 \times 7$ and 3 color \changed{bands}. It includes marker-based normalization before convolution, ReLu activation, and a max-pooling operation with a window of dimension $3 \times 3$. Since it is the single and last convolutional layer, we applied a stride of $4$ in max-pooling and batch normalization to create an input as close as possible to the one of the \changed{MLP} classifier used in the popular VGG-16~\cite{simonyan2014very}. We then vectorized and projected the training images by t-SNE~\cite{maaten2008visualizing} (Figure~\ref{fig:pipeline-xtractor}). Four representative images, only two per class, were chosen by the user to draw markers (Figure~\ref{fig:markers}). These images must come from regions in the projection populated by samples from the same class (either red or green points). Those regions indicate groups of images from the same class in the input feature space. The markers must be placed in regions that best distinguish the coconut and non-coconut classes.}

\changed{Pytorch was chosen to implement FLIM and MLP, being SVM the linear model available in scikit-learn with~${C=0.01}$~\cite{fan2008liblinear, scikit-learn}. We trained MLP for $40$ epochs and the whole network for $60$ epochs with a batch size of $64$. We used a learning rate of $0.0001$, a weight decay of $0.001$, and we updated the learning rate by a factor of $0.1$ at every $5$ epochs after epoch 30. VGG and VGG-FT were trained for $95$ epochs with a batch size of $64$, a learning rate of $0.001$, weight decay of $0.001$, and we updated the learning rate by a factor of $0.1$ at every $5$ epochs after the model achieved accuracy $0.95$.}

\changed{
Table~\ref{tab:results} shows the mean results of this first experiment for three random splits of the dataset into training, validation, and testing sets. We considered better (in bold) the results with non-overlapping intervals of standard deviation around the mean value. The comparison between FLIM+SVM and VGG-FE+SVM indicates that FLIM can produce a feature extractor with a single layer competitive to VGG-FE using 13 layers (i.e., FLIM uses only $0.07\%$ of the number of parameters in VGG-FE). The results of FLIM+MLP, CL+MLP, and FLIM-FT+MLP show that FLIM may dismiss training and fine-tuning based on backpropagation, by directly providing a more effective solution than CL+MLP and FLIM-FT+MLP (H2). This result can also be noticed when comparing  FLIM+MLP with VGG, confirming that FLIM can reduce the number of required training images for the MLP classifier. The comparison between FLIM+MLP and VGG-FT shows competitive results with a top-rated solution (H1).}

\changed{To evaluate H3, we used the validation set to increase the classifiers' number of training images. Note that the marker selection for FLIM was still the same from those four images. To our surprise, as the training set increases, the differences in mean f-score, as obtained among the three splits, favor FLIM+MLP over FLIM-FT+MLP (H2), VGG-FT (H1 and H3), and CL+MLP (H2). The differences between FLIM+MLP and CL+MLP demonstrate that FLIM is superior to backpropagation in filter estimation (H2). Finally, Table II shows the mean results of FLIM+SVM among the three splits as the number of images for marker selection increases, and the training set size remains with 200 images (H4). The results are not conclusive for H4, but the choice of more markers does not deteriorate FLIM's performance. This point deserves further investigation -- i.e., choosing relevant training images and markers for FLIM. }  

\begin{table}[!t]
  \begin{center}
  \begin{tabular}{|l|c|c|c|}
  \hline
   Method & Precision & Recall & F-score \\
  \hline\hline
    FLIM+SVM & 0.856 \textpm 0.011  &  0.831 \textpm 0.019 & 0.838 \textpm 0.017\\ 
    VGG-FE+SVM & 0.855 \textpm 0.001 & 0.816 \textpm 0.007 & 0.826 \textpm 0.006  \\\hline
      
    FLIM+MLP & \textbf{0.863 \textpm 0.002} & \textbf{0.849 \textpm 0.005} & \textbf{0.854 \textpm 0.004}\\
    FLIM-FT+MLP & 0.845 \textpm 0.003 & 0.825 \textpm 0.006 & 0.832 \textpm 0.005 \\
    CL+MLP & 0.822 \textpm 0.011 & 0.805 \textpm 0.013 & 0.811 \textpm 0.012  \\
    VGG & 0.827 \textpm 0.003 & 0.770 \textpm 0.016  &  0.784 \textpm 0.014\\
    VGG-FT & \textbf{0.872 \textpm 0.007} & \textbf{0.844 \textpm 0.015} & \textbf{0.851 \textpm 0.014}  \\
   
  \hline
  \end{tabular}
  \end{center}

\caption{Mean and standard deviation of precision, recall, and f-score for each method: a feature extractor with a classifier.}
  \label{tab:results}
\end{table}

\begin{figure}
  \centering
  \subfloat{\includegraphics[width=.2\linewidth]{figs/qualitative/ours/upper-coco-108.png}
  }
  ~
  \subfloat{\includegraphics[width=.2\linewidth]{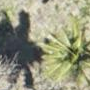}
  }
  ~
  \subfloat{\includegraphics[width=.2\linewidth]{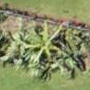}
  }
  \\
  \subfloat{\includegraphics[width=.2\linewidth]{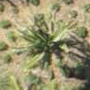}
  }
  ~
  \subfloat{\includegraphics[width=.2\linewidth]{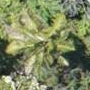}
  }
  ~
  \subfloat{\includegraphics[width=.2\linewidth]{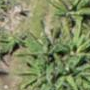}
  }
  \caption{Misclassified images by ``FLIM + MLP'' (first row) and VGG-FT (second row), being the last column with images without coconut trees.}
  \label{fig:ex-classification}
\end{figure}

 In Figure \ref{fig:ex-classification}, we can see examples of misclassified images. In these images, coconut trees appear in different angles, sizes, and shapes, or the boundaries between the coconut tree and the background are tenuous, making it more difficult to identify them. In turn, images that do not contain coconut trees, contain trees that resemble their shape. The network designer could add markers in those images to improve the feature extractor in an active learning loop.

\begin{figure}
  \centering
  \subfloat[\label{fig:vis-input}]{\includegraphics[width=.3\linewidth]{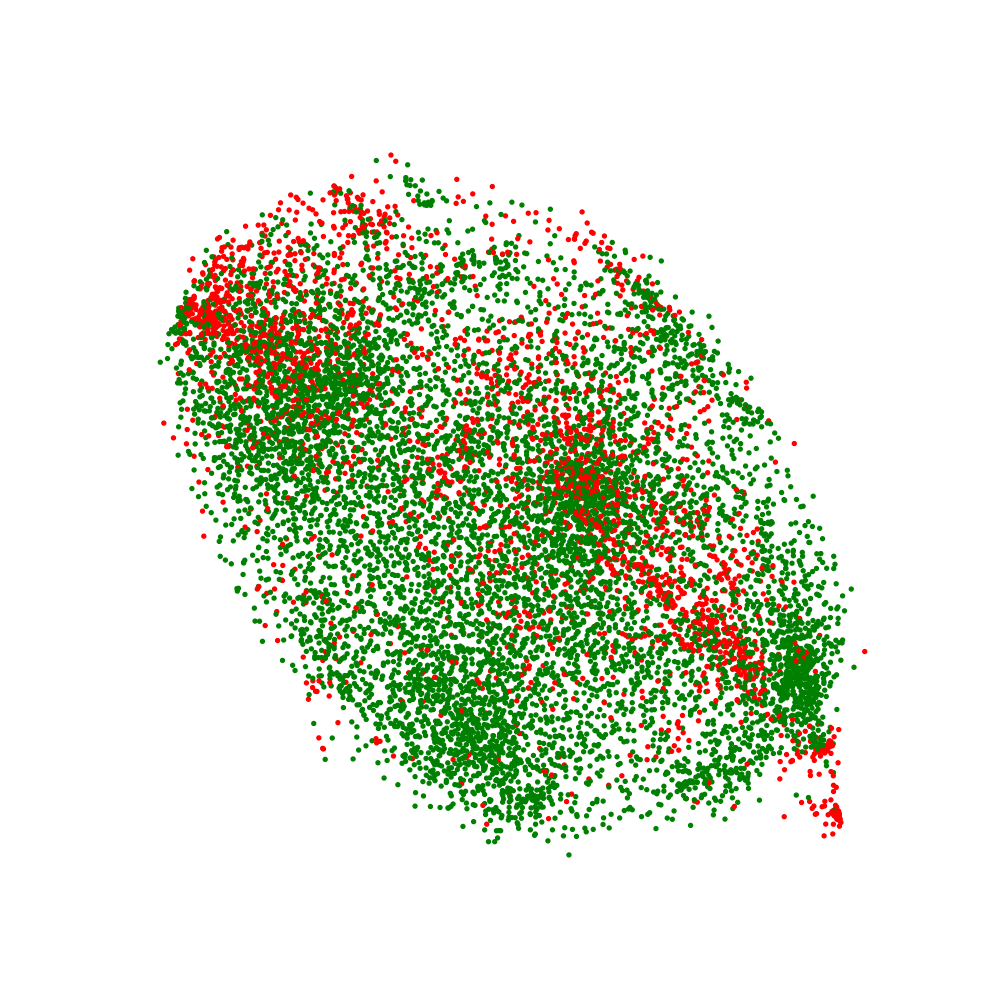}
  }
  ~
  \subfloat[\label{fig:vis-feature-extractor}]{\includegraphics[width=.3\linewidth]{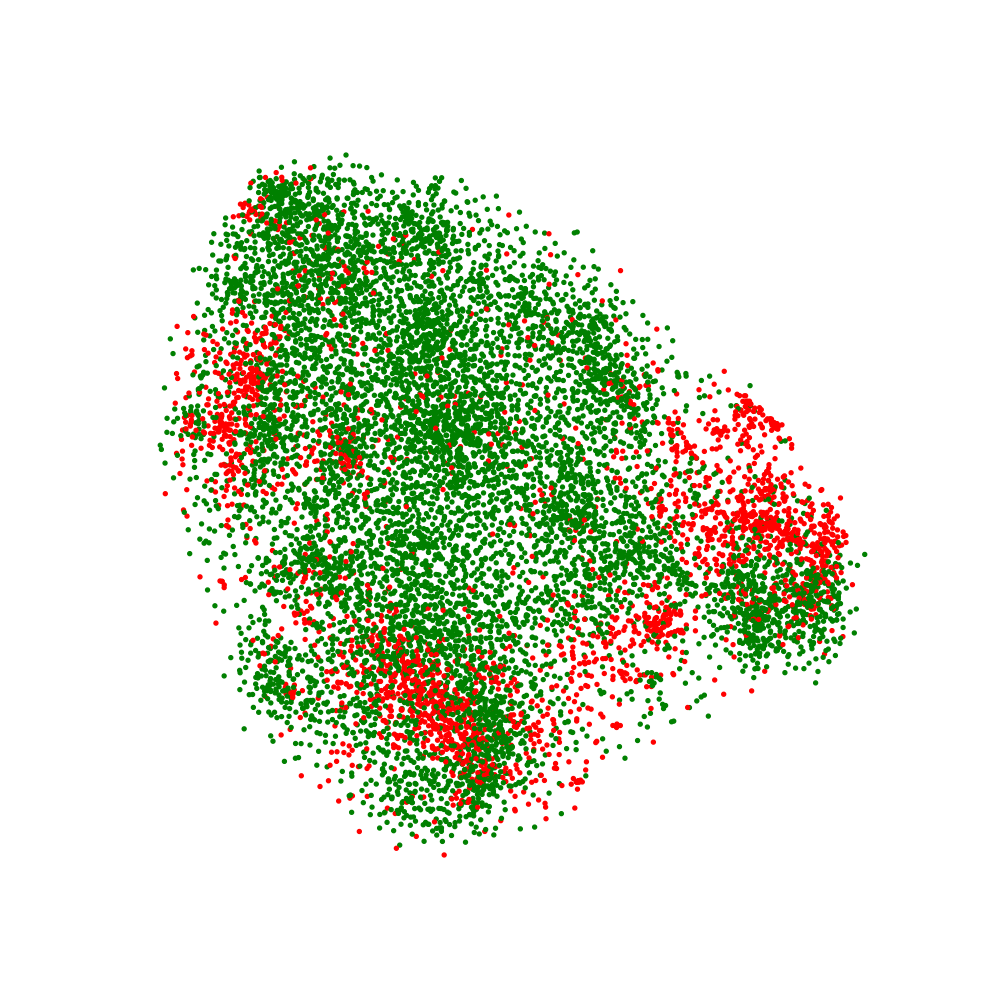}
  }
  ~
  \subfloat[]{\includegraphics[width=.3\linewidth]{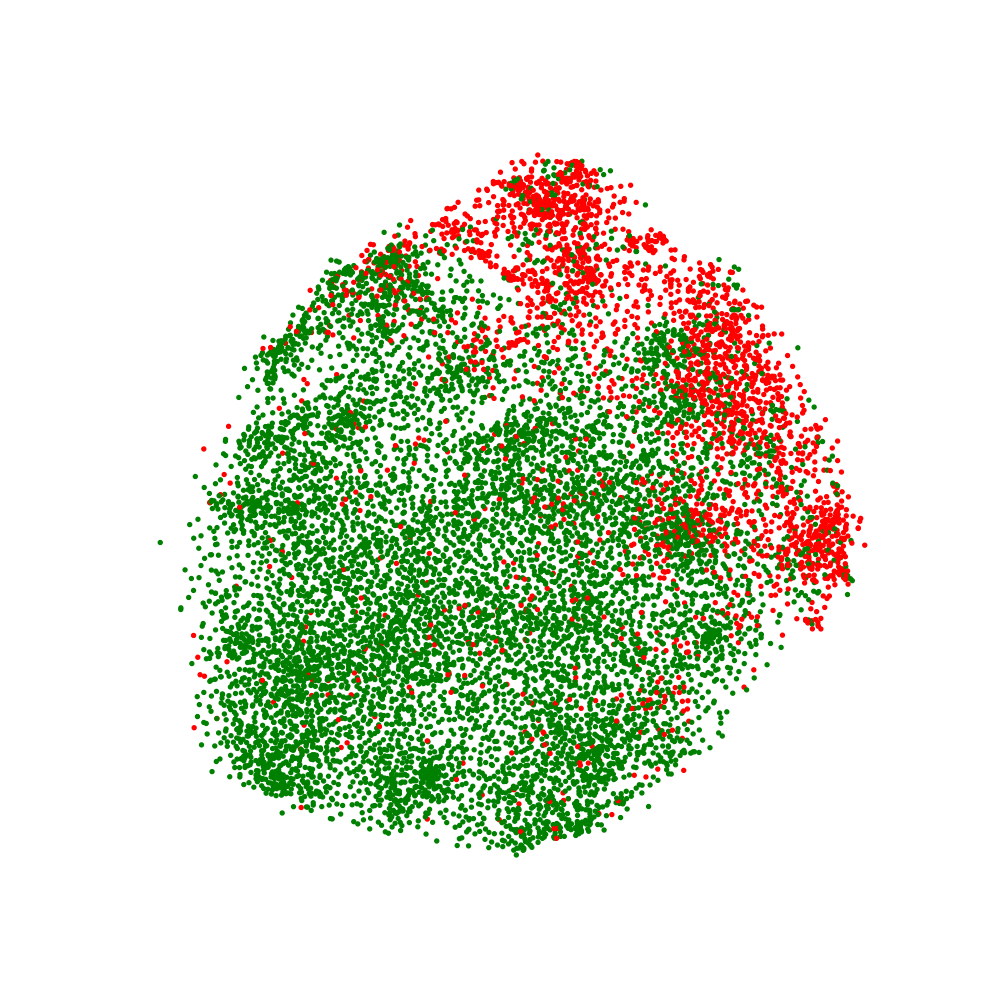}
  \label{fig:vis-classifier}}
  \caption{Projections by t-SNE of the test images at (a) the input layer, (b) the output of the convolutional layer of  FLIM+MLP after batch normalization (Figure~\ref{fig:arch}), and (c) the output of the last hidden layer of the MLP classifier in  FLIM+MLP.}
\end{figure}

As motivated by Rauber et al. \cite{rauber2016visualizing}, we created projections of three stages of the network FLIM+MLP using t-SNE \cite{maaten2008visualizing} to understand how our CNN transforms image spaces along its layers. In Figure \ref{fig:vis-input}, we can see the projection of the test images in the \changed{CIE L*a*b* color space.} Green points are images that contain coconut trees, and red points are images that do not contain coconut trees. The projection shows considerable overlapping between the two classes. The samples of both classes are also dispersed in the projection. In Figure \ref{fig:vis-feature-extractor}, the projection of the output of our feature extractor shows some reduction in sample dispersion. The classes show more concentration in some parts of the projection and reduced overlapping between them. Finally, Figure \ref{fig:vis-classifier} shows the output of the last hidden layer of the MLP classifier in FLIM+MLP. The overlapping between classes and sample dispersion are considerably reduced, explaining our results.

\begin{figure}
  \begin{center}
\begin{tikzpicture}
  \begin{axis}[
    /pgf/number format/set thousands separator = {},
    xlabel = Training set size,
    ylabel = f-score,
    xmin=300, xmax=1200,
    grid=major,
    grid style=dashed,
    xtick={300,450, 600, 750, 900, 1050, 1200},
    legend style={nodes={scale=0.65, transform shape}},
    legend pos=south east
    ]
    \addplot table[x index=0,y index=1,header=false] {data/data.dat};
    \addlegendentry{FLIM+MLP}

    \addplot  table[x index=0,y index=2,header=false] {data/data.dat};
    \addlegendentry{FLIM-FT+MLP}

    \addplot  table[x index=0,y index=3,header=false] {data/data.dat};
    \addlegendentry{VGG-FT}

    \addplot table[x index=0,y index=4,header=false] {data/data.dat};
    \addlegendentry{CL+MLP}

    \legend{FLIM+MLP,FLIM-FT+MLP,VGG-FT,CL+MLP}
  
  \end{axis}
\end{tikzpicture}
\end{center}
\caption{Graph showing the impact of training set size on quantitative results. FLIM+MLP feature extractor was trained only with four images, and the remaining training set was used to train the MLP classifier.}
\end{figure}
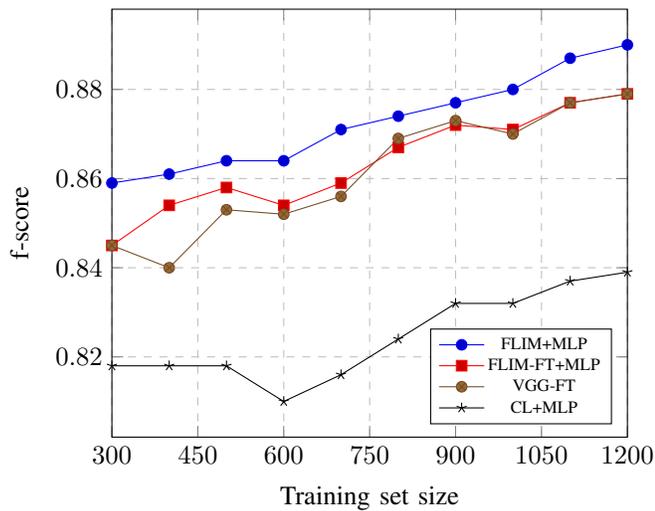

\begin{table}[!t]
    \begin{center}
        \begin{tabular}{|l|c|c|c|}
            \hline
            Method & Precision & Recall & F-score \\
            \hline\hline
            FLIM+SVM (4) & 0.856 \textpm 0.011 &  0.831 \textpm 0.019 &  0.838 \textpm 0.017\\
            FLIM+SVM (8) & 0.857 \textpm 0.005 & 0.838 \textpm 0.010 & 0.844 \textpm 0.009\\
            FLIM+SVM (12) & 0.854 \textpm 0.005 & 0.839 \textpm 0.009 & 0.844 \textpm 0.008\\
            \hline
        \end{tabular}
    \end{center}
    \caption{Mean and standard deviation of precision, recall, and f-score for each method: a feature extractor trained with 8 images and a SVM classifier.}
    \label{tab:results-more-markers}
\end{table}

\section{Conclusion}

We introduced a first feature learning technique, named FLIM, that can estimate effective filter weights for the convolutional layers of a given network architecture from user-drawn markers \changed{in a few images of a training set.} The resulting feature extractor can be used with different classifiers and, when it is used with \changed{an} MLP classifier, the number of required images to train the fully connected layers is reduced. We demonstrated the advantages of FLIM over solutions based on VGG-16 for the binary classification of coconut-tree images. The experiments indicated that FLIM is an effective approach to produce considerably simplified network architectures. By involving the user in the training process, FLIM improves the understanding about CNN models and the user control over the training process. 

We intend to further investigate feature learning from image markers for different applications, \changed{image and marker selection for FLIM,} elaborate methodologies to optimize network architecture, and extend FLIM to estimate weights in fully connected layers.   

\section*{Acknowledgments}

This research was funded by FAPESP (2014/12236-1), CNPq (303808/2018-7), Petróleo Brasileiro S.A. (PETROBRAS) and Agência Nacional do Petróleo, Gás Natural e Biocombustíveis (ANP), Grant Numbers 4600556376 and 4600583791.
\bibliographystyle{IEEEtran}
\bibliography{refs}

\end{document}